\documentclass[letterpaper,journal]{IEEEtran}
\usepackage{amsmath,amsfonts}
\usepackage{algorithmic}
\usepackage{algorithm}
\usepackage{array}
\usepackage{booktabs}
\usepackage{multirow}
\usepackage{makecell}
\usepackage[caption=false,font=normalsize,labelfont=sf,textfont=sf]{subfig}
\usepackage{textcomp}
\usepackage{stfloats}
\usepackage{url}
\usepackage{verbatim}
\usepackage{graphicx}
\usepackage{cite}

\begin{document}

\title{Kinematic-Tactile Aligned Dexterous Retargeting with Contact Wrench Space Alignment}

\author{Xinyuan He, Ziheng Xu, Yueyuan Chen, Yufan Ye, Yuanshuo Tan, Huiming Pan, Jianjun Meng,~\IEEEmembership{Member,~IEEE,} 
Shuo Jiang,~\IEEEmembership{Member,~IEEE,} and
Peter B. Shull,~\IEEEmembership{Member,~IEEE}
\thanks{Manuscript received October 15, 2026; revised Month xx, xxxx; accepted Month x, xxxx. This work was supported by the National Natural Science Foundation of China under Grants W2441018 and 62573321. \textit{(Corresponding author: Shuo Jiang.)}}
\thanks{Xinyuan He, Ziheng Xu, Yueyuan Chen, Yufan Ye, Yuanshuo Tan, Huiming Pan, and Jianjun Meng are with the State Key Laboratory of Mechanical System and Vibration, School of Mechanical Engineering, Shanghai Jiao Tong University, Shanghai 200240, China (e-mail: \{hexinyuan, sjtuxzh0079, yueyuanchen, yeyufan18757557508, tanyuanshuo, panhuiming,  mengjianjunxs008\}@sjtu.edu.cn). Shuo Jiang is with the College of Electronics and Information Engineering, Tongji University, Shanghai 201804, China, and is also with the International Collaboration for Creative Research Teams, Shanghai Jiao Tong University, Shanghai 200240, China (e-mail: jiangshuo@tongji.edu.cn). Peter B. Shull is with the School of Artificial Intelligence, The Chinese University of Hong Kong, Shenzhen, Shenzhen 518172, China (e-mail: pshull@cuhk.edu.cn).}}
\maketitle

\begin{abstract}
Human demonstrations provide a natural and scalable source of data for learning dexterous manipulation policies. 
However, existing retargeting methods primarily transfer human kinematics, while the tactile information underlying physical interaction is rarely preserved across embodiments. 
To address this limitation, we present a hierarchical three-stage framework for kinematic-tactile aligned dexterous retargeting. 
We use object-centric contact wrench space as a physical representation for tactile alignment between human and robot hands. 
The framework first aligns hand kinematics and contact geometry, and then constructs a human wrench support profile from tactile forces and contact geometry. 
Robot fingertip force targets are subsequently optimized to cover the human wrench support and converted into joint position corrections, yielding directly replayable robot hand trajectories. 
We evaluate the framework in simulation and real robot experiments across three pouring conditions. In the real robot experiments, our method achieves a 96.7\% replay success rate and higher wrench transfer fidelity than representative retargeting baselines. 
The framework enables physically grounded transfer of kinematic and tactile information from human demonstrations to robot hands, providing a path toward scalable learning of contact-rich dexterous manipulation.
\end{abstract}

\begin{IEEEkeywords}
Dexterous Manipulation, Learning from Demonstration, Contact Modeling
\end{IEEEkeywords}

\section{Introduction}
\IEEEPARstart{D}{exterous} manipulation has long been challenging due to the complex kinematics of dexterous hands~\cite{rajeswaranLearningComplexDexterous2018}. 
Human demonstrations provide a natural and scalable source of data for learning dexterous manipulation policies, motivating growing efforts to transfer human manipulation skills to anthropomorphic robot hands~\cite{wangDexCapScalablePortable2024}. 
Retargeting plays a critical role in converting human data into robot hand trajectories and has been widely used in both teleoperation and robot-free data collection~\cite{meattiniHumanRobotHand2023,qinDexMVImitationLearning2022}. 
However, existing retargeting methods primarily focus on aligning the kinematics of human and robot hands, while largely overlooking tactile alignment, which plays an important role in reliable and physically consistent manipulation~\cite{chenDexForceExtractingForceinformed2025}. 
Therefore, effective dexterous retargeting should consider both kinematic and tactile alignment to faithfully transfer human manipulation skills across embodiments.

\begin{figure}[!t]
\centering
\includegraphics[width=\columnwidth]{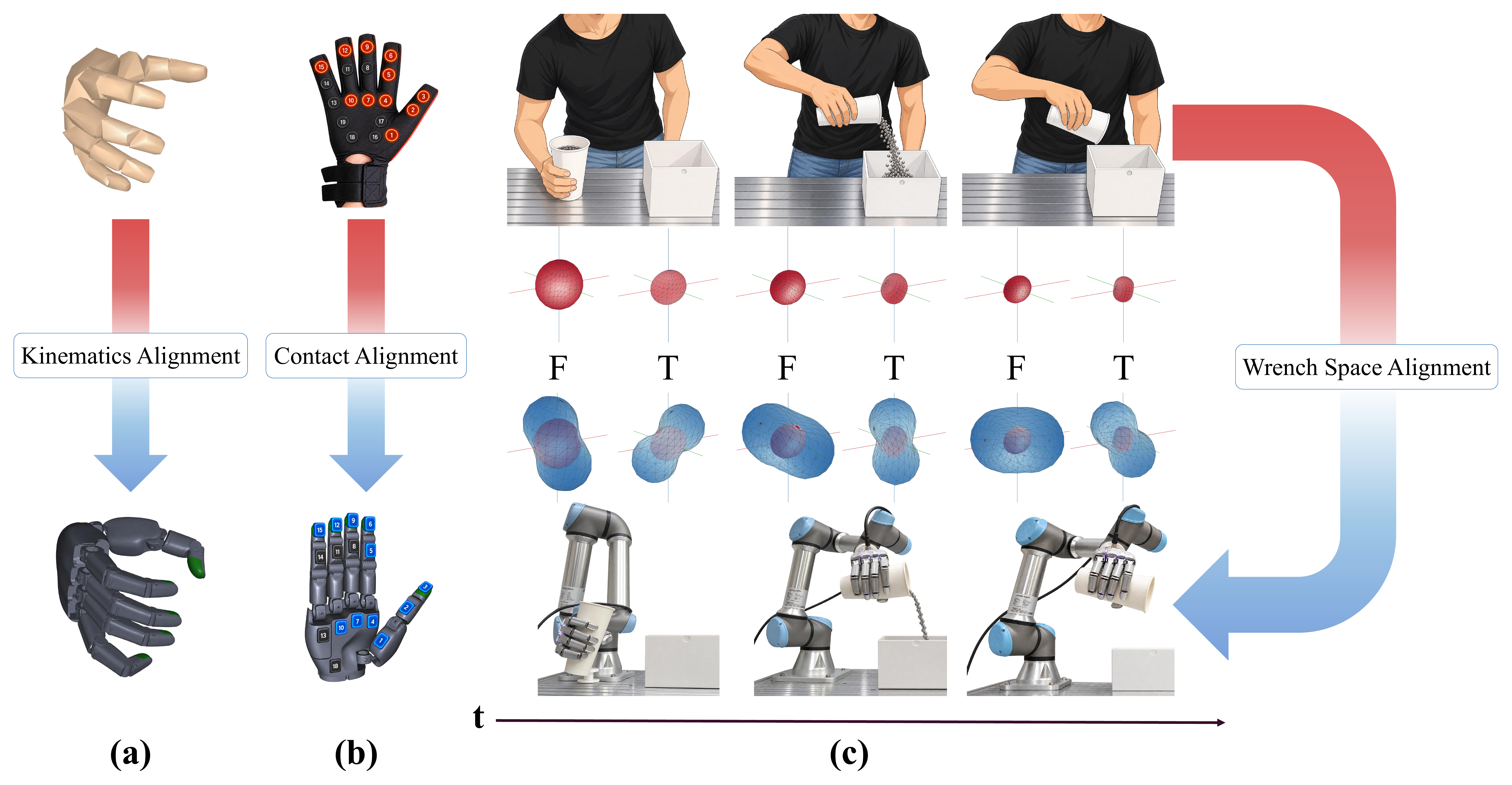}
\caption{Overview of the hierarchical three-stage retargeting framework.
(a) Human-to-robot kinematic alignment transfers reconstructed human motion to
the robot embodiment. (b) Physics-aware contact alignment refines the
hand--object configuration. (c) Object-centric wrench space alignment
optimizes robot normal force targets to cover the human wrench support profile
along the demonstrated manipulation trajectory and converts them into
joint position corrections.}
\label{fig:framework_overview}
\end{figure}

Recent studies have made encouraging progress in functional retargeting, yet the explicit use of human tactile information remains limited. 
Conventional kinematic methods align human and robot hand kinematics through predefined task-space correspondences, such as fingertip positions and relationships between fingers~\cite{handaDexPilotVisionBased2020}. 
More recent interaction-aware methods further preserve contact regions or hand–object interaction structures, improving the consistency of transferred contacts~\cite{lakshmipathyContactTransferDirect2022,wuTopoRetargetInteractionPreservingRetargeting2026}. 
However, these objectives remain geometric and provide no direct constraint on contact forces. 
Modern approaches increasingly employ reinforcement learning to refine retargeted motions through interaction with simulated environments~\cite{liManipTransEfficientDexterous2025a,zhaoDexH2RTaskorientedDexterous2024,mandiDexMachinaFunctionalRetargeting2025}. 
Although these methods can achieve strong performance in simulation, they primarily recover executable motions through policy optimization and task-specific objectives rather than through explicit tactile alignment across embodiments.
Therefore, existing retargeting methods still lack an explicit physical representation to align both kinematic and tactile information for faithful skill transfer across embodiments.

In this letter, we present a hierarchical three-stage retargeting framework that jointly aligns kinematic and tactile information across human and robot embodiments (Fig.~\ref{fig:framework_overview}). 
Our key insight is to use object-centric contact wrench space as a shared physical representation for tactile alignment across different hand embodiments. 
This learning-free formulation enables stable and physically consistent retargeting in real-world manipulation with varying payloads and object deformation.

Specifically, our framework progressively aligns hand kinematics, contact geometry, and object-centric wrench support. 
First, the kinematic alignment stage matches corresponding human and robot task-space vectors frame by frame to generate an initial robot hand trajectory. 
Second, the contact alignment stage uses tactile-activated contact regions to refine the human and robot hand–object configurations at an anchor frame in simulation, and propagates the resulting contact residuals throughout the trajectories. 
Finally, the wrench space alignment stage constructs the human object-centric wrench support profile from contact geometry and tactile forces, and optimizes robot fingertip force targets to cover this profile. 
These force targets are then converted into joint position corrections, yielding the final directly replayable robot hand trajectory.

We evaluate our approach in both simulation and physical experiments on a paper cup pouring task under three payload conditions with different materials and masses. 
The results demonstrate that our framework produces directly replayable robot hand trajectories while preserving human-derived wrench support characteristics across payload conditions. 
In real-world trajectory replay, our method achieves a 96.7\% success rate over 60 trials, outperforming both baselines. 
Our method also achieves higher wrench transfer fidelity than the baselines in both the global and task-relevant wrench spaces.

In summary, our main contributions are:

1. We introduce contact wrench space as a physical representation for tactile alignment in dexterous retargeting.

2. We develop a hierarchical three-stage retargeting framework that progressively aligns kinematic, contact, and wrench information from human demonstrations and produces directly replayable robot hand trajectories.

3. Extensive real robot experiments show that our method achieves higher replay success and wrench transfer fidelity than representative retargeting baselines.
The framework also produces replayable robot demonstrations with human-derived wrench support, providing a practical data source for future contact-rich manipulation learning.

\section{Related Work}

\subsection{Data Collection System}

Demonstrations for dexterous manipulation are acquired through robot
teleoperation, human video, or motion capture. Teleoperation directly yields
trajectories in the robot's configuration space \cite{handaDexPilotVisionBased2020}, whereas video-
and mocap-based pipelines capture natural human hand--object motion
\cite{qinDexMVImitationLearning2022,wangDexCapScalablePortable2024}. Existing
human hand--object datasets provide detailed hand and object kinematics and, in
some cases, geometric contact annotations, but generally lack direct
measurements of time-varying contact forces
\cite{chaoDexYCBBenchmarkCapturing2021,fanARCTICDatasetDexterous2023}.
DexForce extracts force-informed actions from kinesthetic demonstrations;
AET-TacNet infers tactile signals from EMG and hand kinematics; and TactiDex
provides synchronized tactile, kinematic, and object-state data
\cite{chenDexForceExtractingForceinformed2025,
panDecodingHumanTouch2026,niTactiDexRealWorld2026}.
Our acquisition pipeline instead combines synchronized hand--object kinematics
with distributed normal force measurements to construct a time-varying human
wrench support reference for retargeting.

\subsection{Human-to-Robot Dexterous Hand Retargeting}

Kinematics-based hand retargeting aligns joint configurations, fingertip
positions, or relationships between fingers
\cite{qinOneHandMultiple2022,meattiniHumanRobotHand2023,yinGeometricRetargetingPrincipled2025}.
Because these objectives are hand-centric, small alignment errors can alter or
eliminate intended contacts
\cite{lakshmipathyContactTransferDirect2022,wuTopoRetargetInteractionPreservingRetargeting2026}.
Interaction-aware methods transfer contact areas or preserve hand--object
interaction graphs, but their alignment objectives remain primarily geometric
\cite{lakshmipathyContactTransferDirect2022,wuTopoRetargetInteractionPreservingRetargeting2026}.
Contact geometry alone does not determine the contact force distribution or
resulting object wrench
\cite{chenDexForceExtractingForceinformed2025,zhuLearningDexterousManipulation2026}.
Simulation- and policy-based approaches recover executable interactions from
human demonstrations: ManipTrans and DexH2R learn residual corrections to
transferred trajectories; DexMachina tracks object motion through
curriculum-based virtual control; SPIDER refines kinematic references through
physics-based sampling and contact guidance
\cite{liManipTransEfficientDexterous2025a,zhaoDexH2RTaskorientedDexterous2024,mandiDexMachinaFunctionalRetargeting2025,panSPIDERScalablePhysicsInformed2025}.
These approaches improve the physical executability of retargeted motions
through simulation-based refinement or policy learning. In contrast, our
hierarchical framework uses demonstration-derived wrench support as an explicit
retargeting objective alongside kinematic and contact geometry alignment,
yielding directly replayable joint trajectories without policy learning.

\subsection{Wrench Space Representations for Dexterous Manipulation}

Wrench space representations characterize the external forces and torques that
a set of contacts can support and are used to evaluate grasp stability and
synthesize force closure grasps
\cite{ferrariPlanningOptimalGrasps1992,zhuGraspAnalysisSynthesis2004,roaGraspQualityMeasures2015}.
Differentiable force-closure objectives enable diverse grasp synthesis across
arbitrary hand structures \cite{liuSynthesizingDiversePhysically2022}, and
subsequent work accelerates this formulation for large-scale dataset generation
\cite{wangDexGraspNetLargeScaleRobotic2023}. Differentiable
grasp-wrench boundary objectives further support task-oriented pose generation
\cite{chenTaskOrientedDexterousHand2024}, while BODex and Dexonomy incorporate
optimization of contact force feasibility or force-based validation into
scalable grasp synthesis
\cite{chenBODexScalableEfficient2025,chenDexonomySynthesizingAll2025}. These
methods synthesize or validate static hand poses under modeled contact
constraints rather than transfer time-varying interaction mechanics from human
demonstrations. CHORD uses object-centric human--robot contact wrench
consistency to guide reinforcement learning policies
\cite{zhuLearningDexterousManipulation2026}, enabling wrench-level comparison
across embodiments while obtaining robot motions through policy optimization.
In our method, wrench space provides a framewise representation for physical
alignment: a one-sided coverage objective optimizes robot normal force
targets against the human wrench support profile, and these targets are then
converted into joint position corrections.

\begin{figure*}[!t]
\centering
\includegraphics[width=\textwidth]{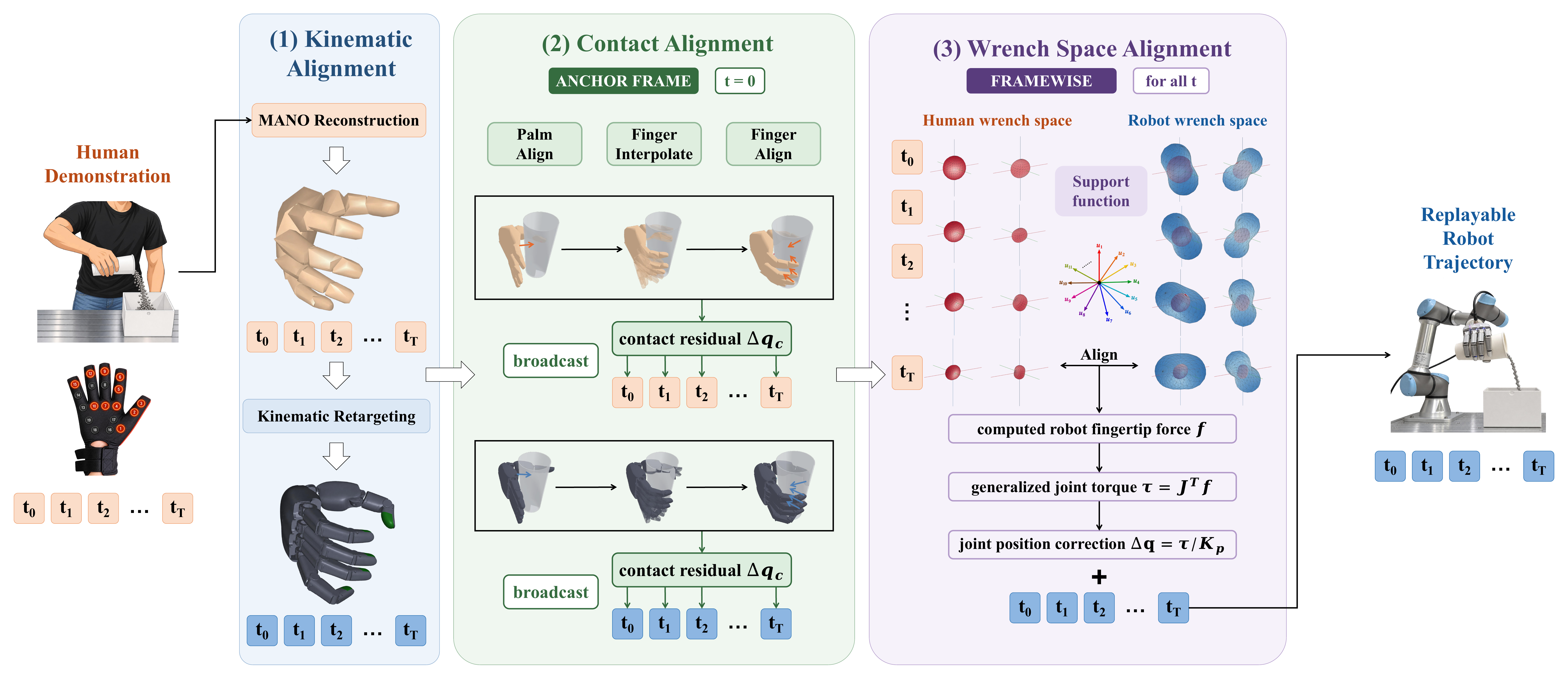}
\caption{Overview of the proposed hierarchical object-centric retargeting
framework. The human demonstration is recorded using motion capture and a
19-channel tactile glove, yielding synchronized hand and object kinematics and
tactile normal force measurements. MoCap-to-MANO reconstruction provides the
human hand sequence. (1) Human-to-robot kinematic alignment produces an initial
robot trajectory. (2) Physics-aware contact alignment locally refines the human
and robot contact configurations in simulation and propagates the contact
residual. (3) Object-centric wrench space alignment optimizes robot normal force
targets to cover the human wrench support profile, converts them into
joint position corrections, and produces a replayable joint position trajectory.}
\label{fig:method_overview}
\end{figure*}

\section{Methods}

Given synchronized trajectories of markers attached to the hand and object, together with tactile
measurements, the proposed framework retargets a human manipulation
demonstration to a dexterous robot hand while transferring its object-centric
wrench support. Following MoCap-to-MANO reconstruction, the framework proceeds through three
hierarchical stages: human-to-robot kinematic alignment, physics-aware contact
alignment, and object-centric wrench space alignment. The final stage converts
optimized robot normal force targets into joint position corrections, yielding
a directly replayable robot joint position trajectory
(Fig.~\ref{fig:method_overview}).

\subsection{MoCap-to-MANO Reconstruction}
\label{sec:demonstration_acquisition}

Synchronized trajectories of markers attached to the hand and object, together with tactile
measurements, provide the kinematic--tactile demonstration.
We represent the human hand using the MANO parametric
model~\cite{romeroEmbodiedHandsModeling2017}. For each subject, we run
HaMeR~\cite{pavlakosReconstructingHands3D2024} on multiple RGB images and take
the elementwise mean of its estimates of the MANO shape vector as
\(\hat{\boldsymbol{\beta}}\). This vector is fixed throughout marker-based
sequence fitting. Each physical marker is manually associated with a MANO
joint and a joint-local offset. The predicted marker position is
\begin{equation}
\hat{\mathbf{y}}_{t,i}
=
\mathbf{p}_{t,k(i)}
+
\mathbf{R}_{t,k(i)}
\Delta\mathbf{p}_i .
\label{eq:mano_marker_prediction}
\end{equation}
where \(\mathbf{p}_{t,k(i)}\) and \(\mathbf{R}_{t,k(i)}\) are the position and
orientation of MANO joint \(k(i)\), respectively, and \(\Delta\mathbf{p}_i\)
is its fixed joint-local marker offset. Let \(\boldsymbol{\xi}_t\) collect the
MANO pose and global wrist parameters. We reconstruct the MANO sequence by
solving
\begin{equation}
\begin{aligned}
\boldsymbol{\xi}^*_{1:T}
=
\arg\min_{\boldsymbol{\xi}_{1:T}}
\big[
&\lambda_m\mathcal{L}_{\mathrm{marker}}
+\lambda_p\mathcal{L}_{\mathrm{pose}}\\
&+\lambda_v\mathcal{L}_{\mathrm{velocity}}
+\lambda_a\mathcal{L}_{\mathrm{acceleration}}
\big].
\end{aligned}
\label{eq:mano_fitting_objective}
\end{equation}
Here, \(\mathcal{L}_{\mathrm{marker}}\) is the confidence-weighted Huber loss
between predicted and measured markers, \(\mathcal{L}_{\mathrm{pose}}\)
penalizes anatomically implausible finger articulations, and \(\mathcal{L}_{\mathrm{velocity}}\) and
\(\mathcal{L}_{\mathrm{acceleration}}\) penalize first- and second-order
temporal differences, respectively. The reconstructed MANO trajectory
\(\{\boldsymbol{\xi}_t^*\}_{t=1}^{T}\) is then used for kinematic retargeting
and human contact reconstruction.

\subsection{Human-to-Robot Kinematic Alignment}
\label{sec:kinematic_retargeting}
We obtain a kinematically aligned robot hand trajectory by minimizing
discrepancies between corresponding human and robot task vectors:
\begin{equation}
\begin{gathered}
\mathbf q_t^{\mathrm{kin}}
=
\underset{\mathbf q}{\arg\min}\;
\sum_{i=1}^{N}
\left\|
\alpha\mathbf v_t^{i}
-\phi_i(\mathbf q)
\right\|_2^2
+\lambda_{\mathrm{s}}
\left\|
\mathbf q-\mathbf q_{t-1}^{\mathrm{kin}}
\right\|_2^2\\
\mathrm{s.t.}\qquad
\mathbf q_{\ell}
\leq
\mathbf q
\leq
\mathbf q_u .
\end{gathered}
\label{eq:kinematic_retargeting_objective}
\end{equation}
Here, \(\mathbf q\) is the robot joint position vector being optimized,
\(N\) is the number of task vector correspondences, and
\(\mathbf v_t^i\) and \(\phi_i(\mathbf q)\) denote the corresponding human
and robot task vectors, respectively. The task vector set comprises
wrist-to-fingertip and inter-fingertip vectors. 
The scale factor \(\alpha\) compensates for the difference in hand size
between the human and robot, while \(\lambda_{\mathrm s}\) promotes temporal smoothness across
successive frames. Joint feasibility is enforced by the bounds
\(\mathbf q_{\ell}\) and \(\mathbf q_u\). At each frame, the bound-constrained problem in
Eq.~\eqref{eq:kinematic_retargeting_objective} is solved using the SLSQP
method. Solving
Eq.~\eqref{eq:kinematic_retargeting_objective} sequentially over all frames
yields the kinematically aligned trajectory
\(\{\mathbf q_t^{\mathrm{kin}}\}_{t=1}^{T}\), which initializes the subsequent
physics-aware contact alignment.

\subsection{Physics-Aware Contact Alignment}
\label{sec:physics_refinement}
Contact consistency is essential for faithfully transferring
the mechanics of hand--object interaction from the human demonstration to the robot hand.
If the intended hand--object contacts are not properly established, the
wrench-guided joint position corrections are diverted toward closing contact
gaps rather than reproducing the desired contact wrench, thereby reducing
transfer fidelity. We therefore enforce contact consistency at an anchor
frame and broadcast the resulting contact residuals across the full
trajectories.

With the object pose held constant, we optimize the wrist and finger poses of
both the MANO and Sharpa Wave hands in MuJoCo. For embodiment
\(X\in\{\mathrm H,\mathrm R\}\), where \(\mathrm H\) and
\(\mathrm R\) denote the human and robot hands, respectively, let \(\mathbf p_i^X\) denote the position of
contact site \(i\) and \(\mathbf p_i^{\mathrm o,X}\) its corresponding nearest
point on the object. The site-level virtual force and its induced generalized
force are
\begin{equation}
\mathbf f_i^{\mathrm v,X}
=
k_f
\left(
\mathbf p_i^{\mathrm o,X}-\mathbf p_i^X
\right),
\qquad
\boldsymbol\tau_i^{\mathrm v,X}
=
\left(\mathbf J_i^X\right)^{\top}
\mathbf f_i^{\mathrm v,X}
\label{eq:virtual_contact_control}
\end{equation}
where \(k_f\) is the virtual force gain and \(\mathbf J_i^X\) is the contact
Jacobian. The position difference, virtual force, and translational rows of
the Jacobian are expressed in the same coordinate frame.
For stage \(s\in\{\mathrm p,\mathrm f\}\), where \(\mathrm p\) and
\(\mathrm f\) denote the palm and finger stages, respectively, let
\(\mathcal C_{\mathrm p}\) and \(\mathcal C_{\mathrm f}\) denote the
corresponding sets of tactile-activated patches. 
These region-specific
sets are disjoint and have different cardinalities. The total generalized
force applied in stage \(s\) is
\begin{equation}
\boldsymbol\tau_s^{\mathrm v,X}
=
\sum_{i\in\mathcal C_s}
\boldsymbol\tau_i^{\mathrm v,X},
\qquad
s\in\{\mathrm p,\mathrm f\}.
\label{eq:stage_virtual_control}
\end{equation}
Only the activated patches in \(\mathcal C_s\) contribute to the corresponding
stage. 

Starting from open-hand poses, we
first activate the palm patches and apply their virtual forces for 200
simulation steps to optimize only the wrist poses. We then interpolate the fingers toward their respective reference poses—reconstructed for MANO and kinematically aligned for the robot.
Finally, we activate the finger patches and apply their virtual forces for another 200 simulation
steps to optimize only the finger joints. A
\(1\,\mathrm{mm}\) MuJoCo contact margin is applied to prevent hand--object
penetration.

We select the first trajectory frame as the anchor, \(t_a=1\). The
refinement described above yields the contact-aligned MANO and robot
states \(\boldsymbol\xi_{t_a}^{\mathrm c}\) and
\(\mathbf q_{t_a}^{\mathrm c}\) at this frame. Their differences from the
reconstructed MANO state \(\boldsymbol\xi_{t_a}^{*}\) and the kinematically
aligned robot state \(\mathbf q_{t_a}^{\mathrm{kin}}\) define fixed contact
residuals, which are broadcast across the full trajectories:
\begin{equation}
\begin{aligned}
\Delta\boldsymbol\xi^{\mathrm c}
&=
\boldsymbol\xi_{t_a}^{\mathrm c}
-
\boldsymbol\xi_{t_a}^{*},
&
\Delta\mathbf q^{\mathrm c}
&=
\mathbf q_{t_a}^{\mathrm c}
-
\mathbf q_{t_a}^{\mathrm{kin}},\\
\boldsymbol\xi_t^{\mathrm c}
&=
\boldsymbol\xi_t^{*}
+
\Delta\boldsymbol\xi^{\mathrm c},
&
\mathbf q_t^{\mathrm c}
&=
\mathbf q_t^{\mathrm{kin}}+\Delta\mathbf q^{\mathrm c},
\quad t=1,\ldots,T.
\end{aligned}
\label{eq:refined_trajectories}
\end{equation}
The resulting contact-aligned MANO and robot trajectories
\(\{\boldsymbol\xi_t^{\mathrm c}\}_{t=1}^{T}\) and
\(\{\mathbf q_t^{\mathrm c}\}_{t=1}^{T}\) provide the human and robot inputs
to the subsequent object-centric wrench space alignment.

\subsection{Object-Centric Wrench Space Alignment}
\label{sec:wrench_alignment}

Given the contact-aligned MANO and robot geometries, we optimize robot
normal force targets to cover the human wrench support profile while holding
contact geometry fixed. In this subsection, contact positions and basis
vectors are expressed in the object frame unless otherwise stated, and
moments are taken about the object center of mass. We omit the frame index
\(t\) and embodiment superscript \(X\in\{\mathrm H,\mathrm R\}\) during the
framewise derivation.

For a hand with \(M\) contacts, let \(\mathbf p_i\in\mathbb R^3\) denote
the position of contact \(i\) and
\(\mathbf Q_i=[\mathbf n_i,\mathbf d_i,\mathbf e_i]\) its orthonormal basis,
where \(\mathbf n_i\) is the inward unit normal and
\(\mathbf d_i,\mathbf e_i\) are unit tangent vectors satisfying
\(\mathbf n_i=\mathbf d_i\times\mathbf e_i\).
For a fixed normal force \(f_{n,i}\geq0\), the Coulomb friction model gives
the admissible force set in local contact coordinates:
\begin{equation}
\mathcal F_i(f_{n,i})
=
\left\{
\mathbf f_i=
\begin{bmatrix}
f_{n,i}\\
f_{d,i}\\
f_{e,i}
\end{bmatrix}
\in\mathbb R^3
\;\middle|\;
f_{d,i}^2+f_{e,i}^2
\leq
\mu_i^2 f_{n,i}^2
\right\},
\label{eq:contact_force_set}
\end{equation}
where \(\mu_i\) is the friction coefficient; we abbreviate this set as
\(\mathcal F_i\). With moment arm
\(\mathbf r_i=\mathbf p_i-\mathbf c^{\mathcal O}\) relative to the object
center of mass \(\mathbf c^{\mathcal O}\), the normalized contact wrench map is
\begin{equation}
\widetilde{\mathbf G}_i
=
\begin{bmatrix}
\mathbf Q_i\\
\rho_{\mathcal O}^{-1}
[\mathbf r_i]_{\times}\mathbf Q_i
\end{bmatrix}
\in\mathbb R^{6\times3}.
\label{eq:normalized_contact_wrench_map}
\end{equation}
Here, \([\mathbf r_i]_{\times}\) is the cross product matrix, and the object
characteristic length \(\rho_{\mathcal O}>0\) normalizes torque relative to
force. The map converts local force components into a wrench expressed in the object frame,
giving \(\widetilde{\mathcal W}_i=\widetilde{\mathbf G}_i\mathcal F_i\).
For fixed normal forces and independent contact force constraints, the
normalized contact wrench space (CWS) is
\begin{equation}
\widetilde{\mathcal W}_C
=
\bigoplus_{i=1}^{M}
\widetilde{\mathcal W}_i,
\label{eq:normalized_cws}
\end{equation}
where \(\bigoplus\) denotes the Minkowski sum, collecting all sums of
admissible individual contact wrenches.

For a nonempty compact set \(\mathcal A\), its support function is
\(h_{\mathcal A}(\mathbf u)=\max_{\mathbf a\in\mathcal A}
\mathbf u^\top\mathbf a\).
We use the support function identities
\(h_{\mathbf L\mathcal A}(\mathbf u)
=h_{\mathcal A}(\mathbf L^\top\mathbf u)\) for a linear map \(\mathbf L\),
and
\(h_{\bigoplus_i\mathcal A_i}(\mathbf u)
=\sum_i h_{\mathcal A_i}(\mathbf u)\) for a finite Minkowski sum.
Let
\(\mathcal U=\{\mathbf u_\ell\}_{\ell=1}^{K}\subset\mathbb S^5\)
denote \(K\) shared support directions sampling the global wrench space in
the object frame.
For contact \(i\) and direction \(\mathbf u_\ell\), define
\begin{equation}
\mathbf v_{\ell i}
=
\widetilde{\mathbf G}_i^{\top}\mathbf u_\ell
=
\begin{bmatrix}
v_{\ell i}^{n}\\
\mathbf v_{\ell i}^{t}
\end{bmatrix},
\label{eq:projected_support_direction}
\end{equation}
where \(v_{\ell i}^{n}\in\mathbb R\) and
\(\mathbf v_{\ell i}^{t}\in\mathbb R^2\) denote the normal and tangential
components, respectively.

Maximizing over the local force set gives the single-contact support:
\begin{equation}
\begin{aligned}
h_{\widetilde{\mathcal W}_i}(\mathbf u_\ell)
&=
h_{\mathcal F_i}
\left(
\widetilde{\mathbf G}_i^{\top}\mathbf u_\ell
\right)\\
&=
f_{n,i}
\left(
v_{\ell i}^{n}
+
\mu_i\|\mathbf v_{\ell i}^{t}\|_2
\right).
\end{aligned}
\label{eq:single_contact_support}
\end{equation}
Let \(b_{\ell i}=v_{\ell i}^{n}
+\mu_i\|\mathbf v_{\ell i}^{t}\|_2\). Because the contact force sets are
independent, the support of their wrench sum is additive:
\begin{equation}
h_{\widetilde{\mathcal W}_C}(\mathbf u_\ell)
=
\sum_{i=1}^{M}
b_{\ell i}f_{n,i}.
\label{eq:multi_contact_support}
\end{equation}

Stacking the support values along all \(K\) directions yields
\begin{equation}
\mathbf h
=
\begin{bmatrix}
h_{\widetilde{\mathcal W}_C}(\mathbf u_1)\\
\vdots\\
h_{\widetilde{\mathcal W}_C}(\mathbf u_K)
\end{bmatrix}
=
\mathbf B\mathbf f_n,
\qquad
[\mathbf B]_{\ell i}=b_{\ell i},
\label{eq:stacked_wrench_support}
\end{equation}
where \(\mathbf f_n=[f_{n,1},\ldots,f_{n,M}]^{\top}\).

Restoring the frame and embodiment indices, the human and robot support
profiles are therefore
\begin{equation}
\mathbf h_t^{\mathrm H}
=
\mathbf B_t^{\mathrm H}\mathbf f_{n,t}^{\mathrm H},
\qquad
\mathbf h_t^{\mathrm R}
=
\mathbf B_t^{\mathrm R}\mathbf f_{n,t}^{\mathrm R}.
\label{eq:human_robot_support}
\end{equation}
For the human hand,
\(\mathbf B_t^{\mathrm H}\in\mathbb R^{K\times M_{\mathrm H}}\)
is constructed from the contact-aligned MANO geometry
\(\boldsymbol\xi_t^{\mathrm c}\), while
\(\mathbf f_{n,t}^{\mathrm H}\) is obtained from the calibrated tactile
measurements, with zero force assigned to inactive contacts. For the robot
hand,
\(\mathbf B_t^{\mathrm R}\in\mathbb R^{K\times M_{\mathrm R}}\)
is constructed from the contact-aligned configuration
\(\mathbf q_t^{\mathrm c}\), while
\(\mathbf f_{n,t}^{\mathrm R}\) contains the normal force targets to be
optimized. In our implementation,
\(M_{\mathrm H}=10\) and \(M_{\mathrm R}=5\).

Rather than enforcing exact support matching, we encourage the robot CWS to
cover the human CWS over the global direction set \(\mathcal U\). For a
candidate robot normal force vector \(\mathbf f\), let
\(\sigma_t=\max(\|\mathbf h_t^{\mathrm H}\|_2/\sqrt K,\epsilon)\), where
\(\epsilon>0\) is a numerical safeguard. We define the one-sided coverage
loss as
\begin{equation}
\mathcal L_{\mathrm{coverage}}(t,\mathbf f)
=
\frac{1}{2K\sigma_t^2}
\left\|
[\mathbf h_t^{\mathrm H}-\mathbf B_t^{\mathrm R}\mathbf f]_+
\right\|_2^2,
\label{eq:cws_alignment_loss}
\end{equation}
where \([x]_+=\max(x,0)\) is applied elementwise. The loss penalizes only
directions in which the robot support is insufficient.

The complete force objective is
\begin{equation}
\begin{aligned}
\mathcal L_{\mathrm{total}}(t,\mathbf f)
= {}&
\lambda_c\mathcal L_{\mathrm{coverage}}(t,\mathbf f)
+\lambda_r\mathcal L_{\mathrm{ratio}}(t,\mathbf f)\\
&+\lambda_t\mathcal L_{\mathrm{temporal}}(t,\mathbf f)
+\lambda_f\mathcal L_{\mathrm{force}}(t,\mathbf f).
\end{aligned}
\label{eq:wrench_guided_force_optimization}
\end{equation}
Here, \(\mathcal L_{\mathrm{ratio}}\) preserves the relative force pattern
measured at the human fingertips, \(\mathcal L_{\mathrm{temporal}}\) promotes
frame-to-frame continuity, and \(\mathcal L_{\mathrm{force}}\) favors smaller
forces among solutions with comparable coverage. The nonnegative
\(\lambda_c\), \(\lambda_r\), \(\lambda_t\), and \(\lambda_f\) weight the
four terms.

We minimize this convex objective frame by frame using a
Nesterov-accelerated projected gradient method. Each iteration projects the
solution onto the feasible set defined by constraints on nonnegativity,
individual finger forces, total force, and variation between frames. Subsequent frames are initialized from the
preceding solution; at \(t=1\), the temporal term and variation constraint are
omitted.
Frames with negligible human support or admissible total force are assigned
zero target force. The resulting
\(\mathbf f_{n,t}^{\mathrm R*}\in\mathbb R^{M_{\mathrm R}}\) contains the five
robot fingertip normal force targets.

With each robot contact normal expressed in the world frame, the desired
fingertip force is
\begin{equation}
\mathbf F_{t,i}^{*}
=
f_{n,t,i}^{\mathrm R*}
\mathbf n_{t,i}^{\mathrm W}.
\label{eq:desired_robot_contact_force}
\end{equation}
The tangential friction component is used to characterize the admissible CWS
but is not explicitly recovered during this conversion. The desired fingertip
forces induce the generalized finger torque
\begin{equation}
\boldsymbol\tau_t^{*}
=
\sum_{i=1}^{M_{\mathrm R}}
\left(\mathbf J_{t,i}^{\mathrm R}\right)^\top
\mathbf F_{t,i}^{*},
\label{eq:wrench_to_joint_torque}
\end{equation}
where \(\mathbf J_{t,i}^{\mathrm R}\) is the translational fingertip Jacobian
evaluated at the contact-aligned robot configuration
\(\mathbf q_t^{\mathrm c}\).

Under a quasi-static proportional position control model,
\(\boldsymbol\tau_t^{*}
\simeq
\mathbf K_p\Delta\mathbf q_t^{\mathrm w}\), yielding
\begin{equation}
\Delta\mathbf q_t^{\mathrm w}
=
\mathbf K_p^{-1}\boldsymbol\tau_t^{*},
\qquad
\mathbf q_t^{\mathrm w}
=
\mathbf q_t^{\mathrm c}
+
\Delta\mathbf q_t^{\mathrm w}.
\label{eq:wrench_guided_position_correction}
\end{equation}
All geometry-dependent quantities are evaluated at
\(\mathbf q_t^{\mathrm c}\), such that
\(\Delta\mathbf q_t^{\mathrm w}\) acts as a local wrench-guided correction.
The resulting joint position trajectory
\(\{\mathbf q_t^{\mathrm w}\}_{t=1}^{T}\)
is finally executed by the robot, while the optimized contact forces serve
only as intermediate physical targets.

\section{Experiments}

The experiments are organized into three parts. We first describe the acquisition of human demonstrations with synchronized kinematic and tactile measurements, which provide the inputs to the proposed retargeting framework. We then conduct real robot experiments as the primary evaluation of trajectory replayability, task success, and human-to-robot wrench transfer. Finally, simulation experiments provide a complementary evaluation of trajectory feasibility and characterize the gap between simulated and real-world execution.

\subsection{Human Demonstration Acquisition}
\label{sec:exp_human_acquisition}

Human demonstrations are collected in the optical motion capture workspace shown in Fig.~\ref{fig:realworld_setup}(a). Hand kinematics are tracked using 21 reflective markers mounted on a tactile glove equipped with 19 sensing channels, while object motion is captured using five reflective markers attached to each manipulated object. The hand motion, object motion, and tactile measurements are synchronized using timestamps to form the aligned kinematic--tactile demonstrations used as input to the proposed framework.

A single operator performs the pouring task under three payload conditions using the same \(22\,\mathrm{oz}\) paper cup as the manipulated object: \(500\,\mathrm{g}\) of rice, \(750\,\mathrm{g}\) of ceramic beads, and \(1500\,\mathrm{g}\) of steel beads. We collect 20 demonstrations for each condition, yielding 60 trajectories in total. The demonstrations are acquired at \(120\,\mathrm{Hz}\), last approximately \(3\)--\(8\,\mathrm{s}\), and comprise about 40,000 synchronized frames in total.

\subsection{Real Robot Experiments}
\label{sec:exp_realworld_setup}

\newcommand{\best}[1]{\textbf{#1}}

\begin{table*}[!t]
\caption{Quantitative real robot evaluation across three pouring payloads.
Success is reported as successful trials out of 20 (percentage), and wrench
metrics are reported as mean $\pm$ standard deviation. The best value for each payload and
metric is shown in bold.}
\label{tab:real_robot_results}
\centering
\scriptsize
\setlength{\tabcolsep}{3.0pt}
\renewcommand{\arraystretch}{1.18}
\resizebox{\textwidth}{!}{
    \begin{tabular}{llccccccc}
        \toprule

        \multirow{2}{*}{Payload}
        &
        \multirow{2}{*}{Method}
        &
        \multirow{2}{*}{
            \makecell{
                Replay success\\
                $\uparrow$
            }
        }
        &
        \multicolumn{3}{c}{Global wrench space}
        &
        \multicolumn{3}{c}{Task-relevant wrench space}
        \\

        \cmidrule(lr){4-6}
        \cmidrule(lr){7-9}

        &
        &
        &
        \makecell{
            Covering rate\\
            $C$ (\%) $\uparrow$
        }
        &
        \makecell{
            Appropriate rate\\
            $A$ (\%) $\uparrow$
        }
        &
        \makecell{
            Margin\\
            $M$ ($\rightarrow 1$)
        }
        &
        \makecell{
            Covering rate\\
            $C$ (\%) $\uparrow$
        }
        &
        \makecell{
            Appropriate rate\\
            $A$ (\%) $\uparrow$
        }
        &
        \makecell{
            Margin\\
            $M$ ($\rightarrow 1$)
        }
        \\

        \midrule

        \multirow{3}{*}{Rice}
        & DexPilot with preload
        & 15/20 (75\%)
        & 32.3 $\pm$ 20.6
        & 29.7 $\pm$ 18.0
        & \best{0.8 $\pm$ 0.4}
        & 45.7 $\pm$ 30.6
        & \best{41.7 $\pm$ 28.9}
        & \best{1.1 $\pm$ 0.7}
        \\

        & ManipTrans
        & 0/20 (0\%)
        & N/A
        & N/A
        & N/A
        & N/A
        & N/A
        & N/A
        \\

        & \textbf{Ours}
        & \best{19/20 (95\%)}
        & \best{52.9 $\pm$ 15.9}
        & \best{40.0 $\pm$ 10.6}
        & 1.6 $\pm$ 1.5
        & \best{60.0 $\pm$ 23.6}
        & 37.6 $\pm$ 28.5
        & 1.9 $\pm$ 0.9
        \\

        \midrule

        \multirow{3}{*}{Ceramic beads}
        & DexPilot with preload
        & 17/20 (85\%)
        & \best{49.8 $\pm$ 23.9}
        & 34.7 $\pm$ 14.6
        & 1.7 $\pm$ 1.7
        & \best{60.5 $\pm$ 29.9}
        & 34.3 $\pm$ 24.3
        & 2.5 $\pm$ 3.0
        \\

        & ManipTrans
        & 0/20 (0\%)
        & N/A
        & N/A
        & N/A
        & N/A
        & N/A
        & N/A
        \\

        & \textbf{Ours}
        & \best{19/20 (95\%)}
        & 47.4 $\pm$ 19.2
        & \best{39.7 $\pm$ 13.0}
        & \best{1.3 $\pm$ 0.7}
        & 57.7 $\pm$ 29.2
        & \best{41.1 $\pm$ 26.7}
        & \best{1.8 $\pm$ 1.6}
        \\

        \midrule

        \multirow{3}{*}{Steel beads}
        & DexPilot with preload
        & 16/20 (80\%)
        & \best{56.2 $\pm$ 21.5}
        & 36.5 $\pm$ 14.6
        & 2.3 $\pm$ 2.8
        & \best{69.6 $\pm$ 30.2}
        & 37.8 $\pm$ 26.7
        & 3.6 $\pm$ 5.0
        \\

        & ManipTrans
        & 0/20 (0\%)
        & N/A
        & N/A
        & N/A
        & N/A
        & N/A
        & N/A
        \\

        & \textbf{Ours}
        & \best{20/20 (100\%)}
        & 53.2 $\pm$ 14.6
        & \best{42.7 $\pm$ 11.6}
        & \best{1.4 $\pm$ 0.7}
        & 67.7 $\pm$ 26.3
        & \best{48.8 $\pm$ 25.0}
        & \best{2.0 $\pm$ 1.3}
        \\

        \bottomrule
    \end{tabular}
}
\end{table*}

\begin{figure}[t]
\centering
\includegraphics[
width=\columnwidth,
trim=800bp 1175bp 975bp 675bp,
clip
]{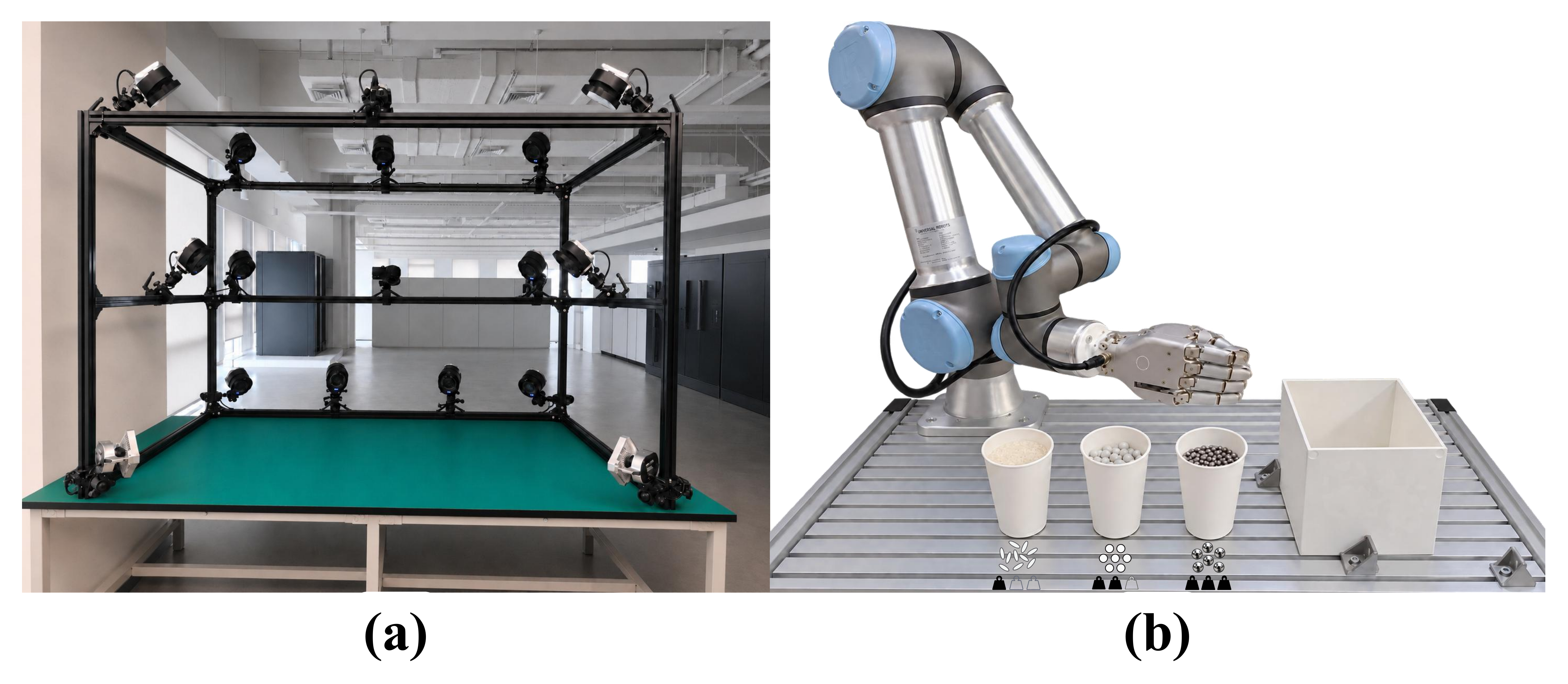}
\caption{Experimental setups. (a) Optical motion capture workspace used to collect human demonstrations. (b) UR7e-mounted Sharpa Wave dexterous hand, three payload conditions, and the fixed receiving container used in the real robot pouring experiments.}
\label{fig:realworld_setup}
\end{figure}

Real-world evaluations are conducted using a Sharpa Wave dexterous hand mounted on a UR7e robot arm, as shown in Fig.~\ref{fig:realworld_setup}(b). During replay, the synchronized arm--hand trajectory targets are updated at \(120\,\mathrm{Hz}\). Both the UR7e and Sharpa Wave controllers operate at \(100\,\mathrm{Hz}\) and apply the latest available targets at each control cycle, while tactile measurements are acquired at \(30\,\mathrm{Hz}\) for each finger. For hardware safety, all trajectories are replayed at \(0.2\times\) the original demonstration speed. We execute the generated joint position trajectories on the physical system without online policy correction to test whether the retargeted interactions can be realized on hardware.

We consider three pouring conditions by loading the same \(22\,\mathrm{oz}\) source cup with \(500\,\mathrm{g}\) of rice, \(750\,\mathrm{g}\) of ceramic beads, or \(1500\,\mathrm{g}\) of steel beads. The receiving container is rigidly fixed to the workbench using metal angle brackets to maintain a consistent target location across trials. The dexterous hand is mounted with a \(51\,\mathrm{mm}\) translational offset along the \(z\)-axis of the UR7e tool center point (TCP) frame. The arm motion is taught separately and synchronized with the retargeted hand trajectory, after which the coupled arm--hand motion executes four phases: grasp and lift, tilt, pour, and hold. To establish a repeatable spatial registration between the robot and the experimental scene, we mount a custom alignment fixture that matches the footprint of the source cup to the UR7e end effector. Using kinesthetic teaching, we register the cup and receiving container locations on the workbench, while metal angle brackets physically constrain their placement across trials. The resulting scene geometry is expressed in the robot base frame through rigid coordinate transformations.

\begin{figure}[!t]
\centering
\includegraphics[width=\columnwidth]{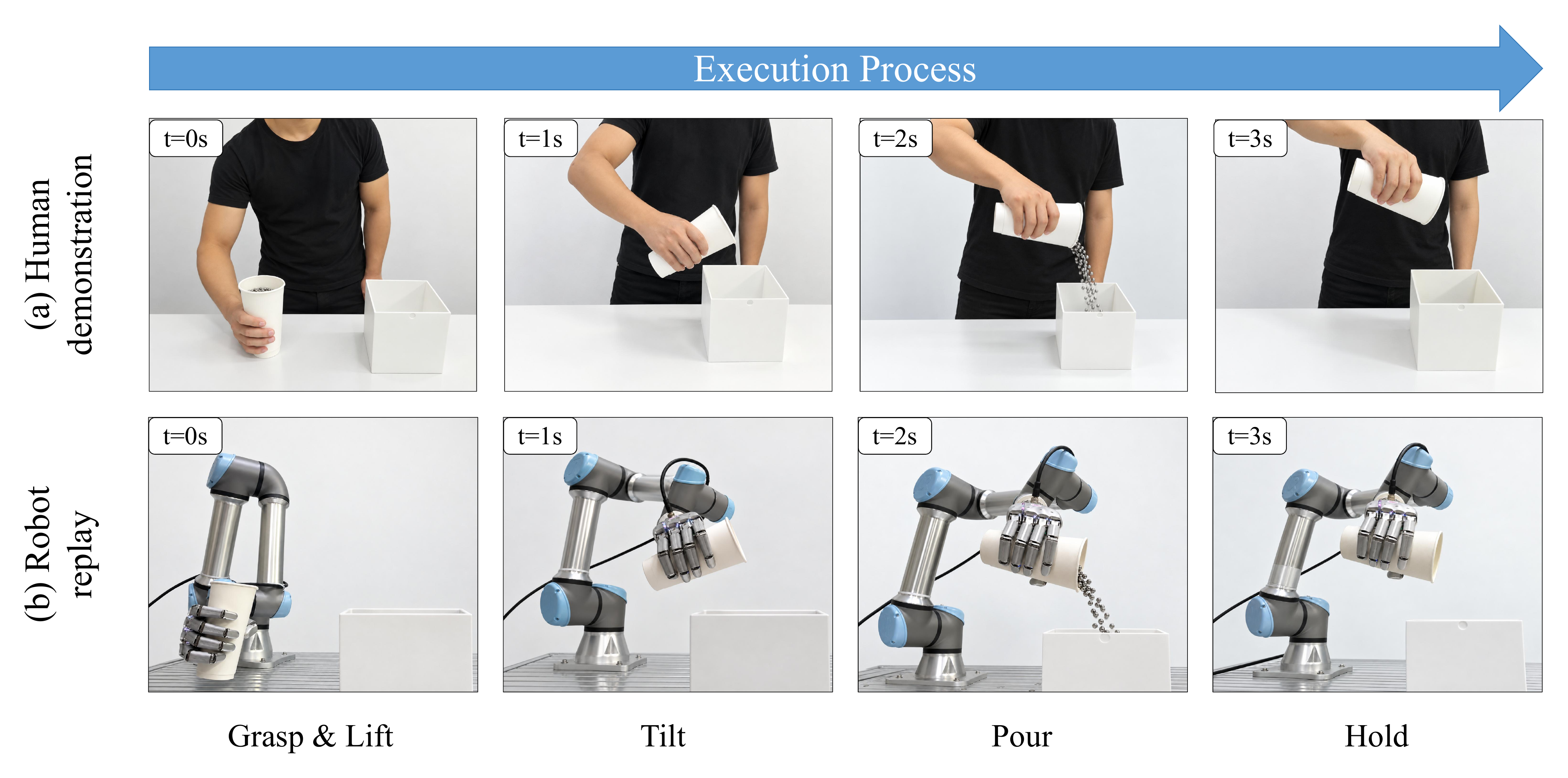}
\caption{Qualitative visualization of a representative pouring sequence.
(a) Human demonstration and (b) the corresponding robot replay across four
task phases: grasp and lift, tilt, pour, and hold.}
\label{fig:realworld_replay_sequence}
\end{figure}

For each replay, the dexterous hand is first moved to the registered cup pose and commanded to execute the first-frame grasp configuration. After establishing the grasp, the arm lifts the cup and moves it to the initial pose of the pouring trajectory, which is transformed with respect to the registered receiving container frame. The synchronized arm--hand trajectory is then replayed from this aligned initial state. We conduct 20 real robot trials for each method--payload pair. A trial is considered successful if the cup remains securely grasped throughout the complete trajectory, exhibits no substantial deformation or loss of grasp stability, and the payload remains contained without unintended spillage. Task performance is reported as the number of successful executions among the 20 trials. Figure~\ref{fig:realworld_replay_sequence} shows representative frames from
a human pouring demonstration and the corresponding robot replay across the
four task phases: grasp and lift, tilt, pour, and hold.

To evaluate interaction fidelity, we compute metrics based on support functions over both the global and task-relevant wrench spaces. The global evaluation queries directions over the normalized six-dimensional wrench space. The task-relevant wrench space evaluation separately considers a force cone centered on the object axis most closely aligned with gravity and a torque cone centered on the axis connecting the fingertips of the human thumb and middle finger; the two sets of metrics are then averaged with equal weight to obtain the task-relevant wrench space macro results. For the real-robot evaluation, the robot support is computed from the
fingertip normal forces measured during replay. At each frame and within each queried direction set, let \(H_k\) and \(R_k\) denote the human and robot support values along direction \(k\), respectively, and define \(V=\{k\mid H_k>\varepsilon\}\) and \(\rho_k=R_k/H_k\) for \(k\in V\), with \(\varepsilon=10^{-6}\). We report three metrics: \textit{(i) covering rate}, \(C=|\{k\mid R_k\geq H_k-\varepsilon\}|/K\), is the fraction of all \(K\) directions in which the robot support meets or exceeds the human reference; \textit{(ii) appropriate rate}, \(A=|\{k\in V\mid 1\leq\rho_k\leq U\}|/|V|\), is the fraction of valid directions providing sufficient but not excessive support, with \(U=3\); and \textit{(iii) margin}, \(M=|V|^{-1}\sum_{k\in V}\rho_k\), is the dimensionless mean support ratio, where values below, equal to, or above one indicate insufficient, matched, or redundant support on average. The reported mean and standard deviation are computed over all available per-frame metric values, with equal weight assigned to each frame.

We compare the complete framework against two baselines. \emph{DexPilot with preload} augments DexPilot kinematic retargeting~\cite{handaDexPilotVisionBased2020} with a fixed contact preload to improve direct physical replay. Starting from the kinematically retargeted hand pose, we project each fingertip tactile patch center onto the cup mesh and define the preload direction from the fingertip toward the corresponding projected point. A \(15\,\mathrm{N}\) preload is then applied along this direction. The resulting fingertip forces are mapped to generalized joint torques and subsequently converted into joint position offsets using the joint stiffness \(K_p\), yielding the final preload-adjusted hand trajectory.

We reproduce \emph{ManipTrans}~\cite{liManipTransEfficientDexterous2025a} following its original method and training protocol. Its two-stage pipeline first pretrains a generalist trajectory imitator to track human hand motions and then fine-tunes a task-specific residual module under interaction constraints. For hardware evaluation, we export the joint trajectories from the simulated
rollouts and replay them open loop using the same physical setup as the other
methods. Table~\ref{tab:real_robot_results} summarizes the quantitative comparison.

Across the three payload conditions, our method succeeds in 58 of 60 trials (96.7\%), whereas DexPilot with preload succeeds in 48 of 60 trials (80.0\%), representing an improvement of 16.7 percentage points. ManipTrans fails all 60 real robot trials. Our method achieves the best value in 11 of the 18 continuous wrench metrics. It improves the global appropriate rate under all three payload conditions and obtains higher task-relevant wrench space appropriate rates and margins closer to unity for both ceramic and steel beads. DexPilot with preload remains stronger on several covering metrics and on several rice-specific measures. The preload-augmented DexPilot baseline serves as a combined ablation of
the two physics-aware stages, assessing their joint benefit over kinematic
retargeting under a fixed \(15\,\mathrm{N}\) contact preload.

Because none of the ManipTrans trajectories yields a valid real robot execution, its wrench transfer metrics are reported as N/A. In particular, wrench fidelity cannot be meaningfully evaluated when the manipulated cup is dropped or substantially deformed, or when the intended pouring sequence is not completed. The difference between its simulation and real-world performance suggests a substantial sim-to-real discrepancy under our deployment conditions, particularly with cup deformation and heavier payloads.

\subsection{Simulation Experiments}
\label{sec:exp_simulation}

Simulation provides a complementary test of trajectory feasibility. We
evaluate our method and DexPilot with preload in MuJoCo and roll out the
complete ManipTrans trajectory in its original Isaac Gym
setting. This protocol retains each method's native simulator.
For each material, 20 distinct trajectories are each rolled out once. A rollout
is successful if the manipulated cup is not dropped throughout the complete
trajectory; accordingly, we report only success rates in
Table~\ref{tab:simulation_results}.

Our method succeeds on all 60 simulated trajectories, while DexPilot with
preload succeeds on 59 of 60 (98.3\%). ManipTrans succeeds on 43 of 60
trajectories (71.7\%): its success rate is 100\% for rice but decreases to 60\%
and 55\% for ceramic and steel beads, respectively. The drop from 71.7\% success in simulation to 0\% on hardware indicates a
substantial sim-to-real gap under our deployment conditions. Because our method and the
preload baseline both approach the simulation ceiling, the real robot
experiments provide the more discriminative evaluation.

\begin{table}[!t]
\caption{Simulation success rates under three payload conditions.}
\label{tab:simulation_results}
\centering
\small
\setlength{\tabcolsep}{3pt}
\renewcommand{\arraystretch}{1.1}
\begin{tabular}{lccc}
\toprule
Material &
\makecell{DexPilot with\\preload} &
ManipTrans &
Ours \\
\midrule
Rice & 20/20 (100\%) & 20/20 (100\%) & 20/20 (100\%) \\
Ceramic beads & 20/20 (100\%) & 12/20 (60\%) & 20/20 (100\%) \\
Steel beads & 19/20 (95\%) & 11/20 (55\%) & 20/20 (100\%) \\
\bottomrule
\end{tabular}
\end{table}

\section{Conclusion}
\label{sec:conclusion}
This letter presented a hierarchical three-stage retargeting framework that jointly aligns kinematic and tactile information across human and robot hands.
The framework uses object-centric contact wrench space as a shared physical representation for tactile alignment, complementing the alignment of hand kinematics and contact geometry.
By converting optimized fingertip force targets into joint position corrections, it produces robot hand trajectories for direct offline replay without online policy correction.
In paper cup pouring experiments under three payload conditions, our method achieved a 96.7\% success rate over 60 real robot trials and improved wrench transfer fidelity in both global and task-relevant wrench spaces.
The resulting robot demonstrations retain human-derived wrench support, providing a practical data source for future contact-rich manipulation learning.
The current evaluation is limited to a single operator, a fixed workspace, and one family of pouring tasks.
Future work will extend the framework to in-hand manipulation and tool use.

%
%
%

\bibliographystyle{IEEEtran}
\bibliography{reference}

@inproceedings{chaoDexYCBBenchmarkCapturing2021,
  title = {{{DexYCB}}: {{A Benchmark}} for {{Capturing Hand Grasping}} of {{Objects}}},
  author = {Chao, Yu-Wei and Yang, Wei and Xiang, Yu and Molchanov, Pavlo and Handa, Ankur and Tremblay, Jonathan and Narang, Yashraj S. and Van Wyk, Karl and Iqbal, Umar and Birchfield, Stan and Kautz, Jan and Fox, Dieter},
  year = {2021},
  booktitle = {2021 {{IEEE}}/{{CVF Conference}} on {{Computer Vision}} and {{Pattern Recognition}} ({{CVPR}})},
  pages = {9040--9049},
  publisher = {IEEE},
  doi = {10.1109/CVPR46437.2021.00893},
  url = {https://ieeexplore.ieee.org/document/9578786/},
}

@inproceedings{chenBODexScalableEfficient2025,
  title = {{{BODex}}: {{Scalable}} and {{Efficient Robotic Dexterous Grasp Synthesis Using Bilevel Optimization}}},
  author = {Chen, Jiayi and Ke, Yubin and Wang, He},
  year = {2025},
  booktitle = {2025 {{IEEE International Conference}} on {{Robotics}} and {{Automation}} ({{ICRA}})},
  pages = {1--8},
  publisher = {IEEE},
  doi = {10.1109/ICRA55743.2025.11127930},
  url = {https://ieeexplore.ieee.org/document/11127930/},
}

@article{chenDexForceExtractingForceinformed2025,
  title = {{{DexForce}}: {{Extracting Force-Informed Actions}} From {{Kinesthetic Demonstrations}} for {{Dexterous Manipulation}}},
  author = {Chen, Claire and Yu, Zhongchun and Choi, Hojung and Cutkosky, Mark and Bohg, Jeannette},
  year = {2025},
  journal = {IEEE Robotics and Automation Letters},
  volume = {10},
  number = {6},
  pages = {6416--6423},
  doi = {10.1109/LRA.2025.3568318},
  url = {https://ieeexplore.ieee.org/document/10993295/},
}

@inproceedings{chenDexonomySynthesizingAll2025,
  title = {Dexonomy: {{Synthesizing All Dexterous Grasp Types}} in a {{Grasp Taxonomy}}},
  author = {Chen, Jiayi and Ke, Yubin and Peng, Lin and Wang, He},
  year = {2025},
  booktitle = {Proceedings of {{Robotics}}: {{Science}} and {{Systems}}},
  publisher = {Robotics: Science and Systems Foundation},
  doi = {10.15607/RSS.2025.XXI.105},
  url = {https://www.roboticsproceedings.org/rss21/p105.html},
}

@inproceedings{chenTaskOrientedDexterousHand2024,
  title = {Task-{{Oriented Dexterous Hand Pose Synthesis Using Differentiable Grasp Wrench Boundary Estimator}}},
  author = {Chen, Jiayi and Chen, Yuxing and Zhang, Jialiang and Wang, He},
  year = {2024},
  booktitle = {2024 {{IEEE}}/{{RSJ International Conference}} on {{Intelligent Robots}} and {{Systems}} ({{IROS}})},
  pages = {5281--5288},
  publisher = {IEEE},
  doi = {10.1109/IROS58592.2024.10802652},
  url = {https://ieeexplore.ieee.org/document/10802652/},
}

@inproceedings{fanARCTICDatasetDexterous2023,
  title = {{{ARCTIC}}: {{A Dataset}} for {{Dexterous Bimanual Hand-Object Manipulation}}},
  author = {Fan, Zicong and Taheri, Omid and Tzionas, Dimitrios and Kocabas, Muhammed and Kaufmann, Manuel and Black, Michael J. and Hilliges, Otmar},
  year = {2023},
  booktitle = {2023 {{IEEE}}/{{CVF Conference}} on {{Computer Vision}} and {{Pattern Recognition}} ({{CVPR}})},
  pages = {12943--12954},
  publisher = {IEEE},
  doi = {10.1109/CVPR52729.2023.01244},
  url = {https://ieeexplore.ieee.org/document/10203858/},
}

@inproceedings{ferrariPlanningOptimalGrasps1992,
  title = {Planning Optimal Grasps},
  booktitle = {Proceedings 1992 {{IEEE International Conference}} on {{Robotics}} and {{Automation}}},
  author = {Ferrari, C. and Canny, J.},
  year = {1992},
  pages = {2290--2295},
  publisher = {IEEE Comput. Soc. Press},
  address = {Nice, France},
  doi = {10.1109/ROBOT.1992.219918},
  url = {http://ieeexplore.ieee.org/document/219918/},
  eventtitle = {1992 {{IEEE International Conference}} on {{Robotics}} and {{Automation}}},
  isbn = {978-0-8186-2720-0},
}

@inproceedings{handaDexPilotVisionBased2020,
  title = {{{DexPilot}}: {{Vision-Based Teleoperation}} of {{Dexterous Robotic Hand-Arm System}}},
  author = {Handa, Abhinav and Van Wyk, Karl and Yang, Wei and Liang, Jacky and Chao, Yu-Wei and Wan, Qian and Birchfield, Stan and Ratliff, Nathan and Fox, Dieter},
  year = {2020},
  booktitle = {2020 {{IEEE International Conference}} on {{Robotics}} and {{Automation}} ({{ICRA}})},
  pages = {9164--9170},
  publisher = {IEEE},
  doi = {10.1109/ICRA40945.2020.9197124},
  url = {https://ieeexplore.ieee.org/document/9197124/},
}

@inproceedings{lakshmipathyContactTransferDirect2022,
  title = {Contact {{Transfer}}: {{A Direct}}, {{User-Driven Method}} for {{Human}} to {{Robot Transfer}} of {{Grasps}} and {{Manipulations}}},
  author = {Lakshmipathy, Arjun and Bauer, Dominik and Bauer, Cornelia and Pollard, Nancy S.},
  year = {2022},
  booktitle = {2022 {{International Conference}} on {{Robotics}} and {{Automation}} ({{ICRA}})},
  pages = {6195--6201},
  publisher = {IEEE},
  doi = {10.1109/ICRA46639.2022.9811739},
  url = {https://ieeexplore.ieee.org/document/9811739/},
}

@inproceedings{liManipTransEfficientDexterous2025a,
  title = {{{ManipTrans}}: {{Efficient Dexterous Bimanual Manipulation Transfer}} via {{Residual Learning}}},
  author = {Li, Kailin and Li, Puhao and Liu, Tengyu and Li, Yuyang and Huang, Siyuan},
  year = {2025},
  booktitle = {Proceedings of the {{IEEE}}/{{CVF Conference}} on {{Computer Vision}} and {{Pattern Recognition}} ({{CVPR}})},
  pages = {6991--7003},
  publisher = {IEEE},
  doi = {10.1109/CVPR52734.2025.00656},
  url = {https://ieeexplore.ieee.org/document/11094147/},
}

@article{liuSynthesizingDiversePhysically2022,
  title = {Synthesizing {{Diverse}} and {{Physically Stable Grasps}} with {{Arbitrary Hand Structures}} Using {{Differentiable Force Closure Estimator}}},
  author = {Liu, Tengyu and Liu, Zeyu and Jiao, Ziyuan and Zhu, Yixin and Zhu, Song-Chun},
  year = {2022},
  journal = {IEEE Robotics and Automation Letters},
  volume = {7},
  number = {1},
  pages = {470--477},
  publisher = {IEEE},
  doi = {10.1109/LRA.2021.3129138},
  url = {https://ieeexplore.ieee.org/document/9619920/},
}

@inproceedings{mandiDexMachinaFunctionalRetargeting2025,
  title = {{{DexMachina}}: {{Functional Retargeting}} for {{Bimanual Dexterous Manipulation}}},
  author = {Zhao, Mandi and Hou, Yifan and Fox, Dieter and Narang, Yashraj and Mandlekar, Ajay and Song, Shuran},
  year = {2026},
  booktitle = {Proceedings of the 43rd International Conference on Machine Learning},
  series = {Proceedings of Machine Learning Research},
  volume = {306},
  pages = {86198--86213},
  publisher = {PMLR},
  url = {https://proceedings.mlr.press/v306/mandi26a.html},
}

@article{meattiniHumanRobotHand2023,
  title = {Human to {{Robot Hand Motion Mapping Methods}}: {{Review}} and {{Classification}}},
  author = {Meattini, Roberto and Suárez, Raúl and Palli, Gianluca and Melchiorri, Claudio},
  year = {2023},
  journal = {IEEE Transactions on Robotics},
  volume = {39},
  number = {2},
  pages = {842--861},
  issn = {1552-3098, 1941-0468},
  doi = {10.1109/TRO.2022.3205510},
  url = {https://ieeexplore.ieee.org/document/9896989/},
}

@article{panDecodingHumanTouch2026,
  title = {Decoding {{Human Touch Noninvasively}}: {{Tactile Inference From EMG}} and {{Kinematics Using AET-TacNet}}},
  author = {Pan, Huiming and Zhu, Kezhe and Li, Dongxuan and Chen, Yueyuan and Jiang, Shuo and He, Bin and Shull, Peter B.},
  year = {2026},
  journal = {IEEE Transactions on Industrial Informatics},
  volume = {22},
  number = {6},
  pages = {5117--5128},
  doi = {10.1109/TII.2026.3665344},
  url = {https://doi.org/10.1109/TII.2026.3665344},
}

@misc{panSPIDERScalablePhysicsInformed2025,
  title = {{{SPIDER}}: {{Scalable Physics-Informed Dexterous Retargeting}}},
  author = {Pan, Chaoyi and Wang, Changhao and Qi, Haozhi and Liu, Zixi and Bharadhwaj, Homanga and Sharma, Akash and Wu, Tingfan and Shi, Guanya and Malik, Jitendra and Hogan, Francois},
  year = {2026},
  note = {arXiv:2511.09484; accepted to the 2026 IEEE/RSJ International Conference on Intelligent Robots and Systems (IROS 2026)},
  doi = {10.48550/arXiv.2511.09484},
  url = {https://arxiv.org/abs/2511.09484},
}

@inproceedings{pavlakosReconstructingHands3D2024,
  title = {Reconstructing Hands in 3{D} with Transformers},
  author = {Pavlakos, Georgios and Shan, Dandan and Radosavovic, Ilija and Kanazawa, Angjoo and Fouhey, David and Malik, Jitendra},
  year = {2024},
  booktitle = {2024 {{IEEE}}/{{CVF Conference}} on {{Computer Vision}} and {{Pattern Recognition}} ({{CVPR}})},
  pages = {9826--9836},
  publisher = {IEEE},
  doi = {10.1109/CVPR52733.2024.00938},
  url = {https://ieeexplore.ieee.org/document/10655481/},
}

@inproceedings{qinDexMVImitationLearning2022,
  title = {{{DexMV}}: {{Imitation Learning}} for {{Dexterous Manipulation}} from {{Human Videos}}},
  author = {Qin, Yuzhe and Wu, Yueh-Hua and Liu, Shaowei and Jiang, Hanwen and Yang, Ruihan and Fu, Yang and Wang, Xiaolong},
  year = {2022},
  booktitle = {Computer Vision -- {{ECCV}} 2022},
  series = {Lecture Notes in Computer Science},
  volume = {13699},
  pages = {570--587},
  publisher = {Springer Nature Switzerland},
  address = {Cham},
  doi = {10.1007/978-3-031-19842-7_33},
  url = {https://link.springer.com/chapter/10.1007/978-3-031-19842-7_33},
}

@article{qinOneHandMultiple2022,
  title = {From {{One Hand}} to {{Multiple Hands}}: {{Imitation Learning}} for {{Dexterous Manipulation From Single-Camera Teleoperation}}},
  author = {Qin, Yuzhe and Su, Hao and Wang, Xiaolong},
  year = {2022},
  journal = {IEEE Robotics and Automation Letters},
  volume = {7},
  number = {4},
  pages = {10873--10881},
  issn = {2377-3766, 2377-3774},
  doi = {10.1109/LRA.2022.3196104},
  url = {https://ieeexplore.ieee.org/document/9849105/},
}

@inproceedings{rajeswaranLearningComplexDexterous2018,
  title = {Learning {{Complex Dexterous Manipulation}} with {{Deep Reinforcement Learning}} and {{Demonstrations}}},
  author = {Rajeswaran, Aravind and Kumar, Vikash and Gupta, Abhishek and Vezzani, Giulia and Schulman, John and Todorov, Emanuel and Levine, Sergey},
  year = {2018},
  booktitle = {Proceedings of {{Robotics}}: {{Science}} and {{Systems}}},
  address = {Pittsburgh, Pennsylvania},
  publisher = {Robotics: Science and Systems Foundation},
  doi = {10.15607/RSS.2018.XIV.049},
  url = {https://www.roboticsproceedings.org/rss14/p49.html},
}

@article{roaGraspQualityMeasures2015,
  title = {Grasp Quality Measures: Review and Performance},
  author = {Roa, Máximo A. and Suárez, Raúl},
  year = {2015},
  journal = {Autonomous Robots},
  volume = {38},
  number = {1},
  pages = {65--88},
  issn = {0929-5593, 1573-7527},
  doi = {10.1007/s10514-014-9402-3},
  url = {http://link.springer.com/10.1007/s10514-014-9402-3},
}

@article{romeroEmbodiedHandsModeling2017,
  title = {Embodied Hands: Modeling and Capturing Hands and Bodies Together},
  author = {Romero, Javier and Tzionas, Dimitrios and Black, Michael J.},
  year = {2017},
  journal = {ACM Transactions on Graphics},
  volume = {36},
  number = {6},
  pages = {245:1--245:17},
  doi = {10.1145/3130800.3130883},
  url = {https://dl.acm.org/doi/10.1145/3130800.3130883},
}

@inproceedings{wangDexCapScalablePortable2024,
  title = {{{DexCap}}: {{Scalable}} and {{Portable Mocap Data Collection System}} for {{Dexterous Manipulation}}},
  author = {Wang, Chen and Shi, Haochen and Wang, Weizhuo and Zhang, Ruohan and Fei-Fei, Li and Liu, C. Karen},
  year = {2024},
  booktitle = {Proceedings of {{Robotics}}: {{Science}} and {{Systems}}},
  publisher = {Robotics: Science and Systems Foundation},
  doi = {10.15607/RSS.2024.XX.043},
  url = {https://www.roboticsproceedings.org/rss20/p043.html},
}

@inproceedings{wangDexGraspNetLargeScaleRobotic2023,
  title = {{{DexGraspNet}}: {{A Large-Scale Robotic Dexterous Grasp Dataset}} for {{General Objects Based}} on {{Simulation}}},
  author = {Wang, Ruicheng and Zhang, Jialiang and Chen, Jiayi and Xu, Yinzhen and Li, Puhao and Liu, Tengyu and Wang, He},
  year = {2023},
  booktitle = {2023 {{IEEE International Conference}} on {{Robotics}} and {{Automation}} ({{ICRA}})},
  pages = {11359--11366},
  publisher = {IEEE},
  doi = {10.1109/ICRA48891.2023.10160982},
  url = {https://ieeexplore.ieee.org/document/10160982/},
}

@misc{wuTopoRetargetInteractionPreservingRetargeting2026,
  title = {{{TopoRetarget}}: {{Interaction-Preserving Retargeting}} for {{Dexterous Manipulation}}},
  author = {Wu, Jielin and Yao, Shenzhe and He, Guanqi and Liu, Xiaohan and Zeng, Zhaoqing and Jiang, Xiangrui and Yang, Han and Zhang, Wentao and Zhao, Hang},
  year = {2026},
  note = {arXiv:2606.16272; accepted to the Conference on Robot Learning (CoRL 2026)},
  doi = {10.48550/arXiv.2606.16272},
  url = {https://arxiv.org/abs/2606.16272},
}

@inproceedings{yinGeometricRetargetingPrincipled2025,
  title = {Geometric {{Retargeting}}: {{A Principled}}, {{Ultrafast Neural Hand Retargeting Algorithm}}},
  booktitle = {2025 {{IEEE}}/{{RSJ International Conference}} on {{Intelligent Robots}} and {{Systems}} ({{IROS}})},
  author = {Yin, Zhao-Heng and Wang, Changhao and Pineda, Luis and Bodduluri, Krishna and Wu, Tingfan and Abbeel, Pieter and Mukadam, Mustafa},
  year = {2025},
  pages = {17376--17382},
  publisher = {IEEE},
  address = {Hangzhou, China},
  doi = {10.1109/IROS60139.2025.11247700},
  url = {https://ieeexplore.ieee.org/document/11247700/},
  eventtitle = {2025 {{IEEE}}/{{RSJ International Conference}} on {{Intelligent Robots}} and {{Systems}} ({{IROS}})},
  isbn = {979-8-3315-4393-8},
}

@article{zhaoDexH2RTaskorientedDexterous2024,
  title = {{{DexH2R}}: {{Task-Oriented Dexterous Manipulation}} From {{Human}} to {{Robots}}},
  author = {Zhao, Shuqi and Zhu, Xinghao and Chen, Yuxin and Li, Chenran and Xie, Yichen and Zhang, Xiang and Ding, Mingyu and Tomizuka, Masayoshi},
  year = {2026},
  journal = {IEEE/ASME Transactions on Mechatronics},
  volume = {31},
  number = {3},
  pages = {3202--3213},
  doi = {10.1109/TMECH.2025.3641164},
  url = {https://ieeexplore.ieee.org/document/11316901/},
}

@article{zhuGraspAnalysisSynthesis2004,
  title = {Grasp Analysis and Synthesis Based on a New Quantitative Measure},
  author = {Zhu, Xiangyang and Ding, Han and Wang, Jun},
  year = {2004},
  journal = {Robotics and Automation, IEEE Transactions on},
  volume = {19},
  pages = {942--953},
  doi = {10.1109/TRA.2003.819604},
}

@misc{zhuLearningDexterousManipulation2026,
  title = {Learning {{Dexterous Manipulation Using Contact Wrench Guidance From Human Demonstration}}},
  author = {Zhu, Xinghao and Liu, Zixi and Jain, Shalin and Li, Chenran and Noori, Milad and Lin, Michael Andres and Zhao, Huihua and Welsh, John and Verghese, Mrinal and Liu, Wei and Wang, Tingwu and Da, Xingye and Luo, Zhengyi and Kulkarni, Vishal and Bhatti, Naema and Zhu, Yuke and Fan, Linxi and Wen, Bowen and Xu, Danfei and Pouya, Soha and Chang, Yan},
  year = {2026},
  note = {arXiv:2607.00033; accepted to the Conference on Robot Learning (CoRL) 2026},
  doi = {10.48550/arXiv.2607.00033},
  url = {https://arxiv.org/abs/2607.00033},
}

@misc{niTactiDexRealWorld2026,
  title = {{{TactiDex}}: {{A Real-World Tactile-Guided Benchmark}} for {{Human-Like Dexterous Manipulation}}},
  author = {Ni, Suting and Zhang, Hanbing and Wei, Zhenyu and Chen, Guo and Zhang, Chixuan and Shi, Ye and Wang, Jingya},
  year = {2026},
  note = {arXiv:2607.09190; accepted to the 34th ACM International Conference on Multimedia (ACM MM 2026)},
  doi = {10.48550/arXiv.2607.09190},
  url = {https://arxiv.org/abs/2607.09190},
}

\end{document}